\documentclass[]{interact}

\usepackage{epstopdf}
\usepackage[caption=false]{subfig}
\usepackage{multirow}

\usepackage[numbers,sort&compress]{natbib}
\bibpunct[, ]{[}{]}{,}{n}{,}{,}
\renewcommand\bibfont{\fontsize{10}{12}\selectfont}
\makeatletter
\def\NAT@def@citea{\def\@citea{\NAT@separator}}
\makeatother

\theoremstyle{plain}

\theoremstyle{definition}

\theoremstyle{remark}

\begin{document}

\articletype{Full Paper}

\title{Motion Generation for Food Topping Challenge 2024: \\ Serving Salmon Roe Bowl and Picking Fried Chicken}

\author{
\name{
    Koki Inami\textsuperscript{a}\thanks{CONTACT Koki Inami. Email: inami.koki.qy@alumni.tsukuba.ac.jp}, 
    Masashi Konosu\textsuperscript{a}, 
    Koki Yamane\textsuperscript{a}, 
    Nozomu Masuya\textsuperscript{a},
    Yunhan Li\textsuperscript{a},
    Yu-Han Shu\textsuperscript{a}, 
    Hiroshi Sato\textsuperscript{a}, 
    Shinnosuke Homma\textsuperscript{a}, and 
    Sho Sakaino\textsuperscript{b}}
\affil{
\textsuperscript{a}Intelligent and Mechanical Interaction Systems, Degree Programs in Systems and Information and Engineering, Graduate School of Science and Technology, University of Tsukuba, Japan;
\textsuperscript{b}Department of Intelligent Interaction Technologies, Institute of Systems and Information Engineering, University of Tsukuba, Japan;}
}

\maketitle

\begin{abstract}
Although robots have been introduced in many industries, food production robots are yet to be widely employed because the food industry requires not only delicate movements to handle food but also complex movements that adapt to the environment. Force control is important for handling delicate objects such as food. In addition, achieving complex movements is possible by making robot motions based on human teachings. Four-channel bilateral control is proposed, which enables the simultaneous teaching of position and force information. Moreover, methods have been developed to reproduce motions obtained through human teachings and generate adaptive motions using learning. We demonstrated the effectiveness of these methods for food handling tasks in the Food Topping Challenge at the 2024 IEEE International Conference on Robotics and Automation (ICRA 2024). For the task of serving salmon roe on rice, we achieved the best performance because of the high reproducibility and quick motion of the proposed method. Further, for the task of picking fried chicken, we successfully picked the most pieces of fried chicken among all participating teams. This paper describes the implementation and performance of these methods.

\end{abstract}


\section{Introduction}

Robot automation has progressed in recent years, especially in the manufacturing industry because robots excel at precise repetitive movements in a well-equipped environment. However, their use in the food industry has been limited because of the complexity of the work and the need to generate movements that adapt to the work environment. In addition, food products are difficult to handle with robots because of the variety, fragility, and irregularity of the object. In addition to these factors, high throughput is required because the unit cost of food products is lower than that of manufacturing industrial products.

Many learning-based methods have been proposed to overcome these challenges. One such method is reinforcement learning~\cite{DeepRein_Kai2017}, wherein the agent learns autonomously by designing a reward function.
This has been used successfully in games such as Go, Chess, and Shogi~\cite{MasteringGo_silver2017, MasteringChessShogi_silver2017}. 
Several research studies have applied reinforcement learning to robots~\cite{GoogleRL_levine2018, RLopenAI_andrychowicz2020}; however, it requires many trials for learning. Therefore, creating models that adapt to various environments is very time consuming. In addition, with robots that have physical realities, challenges such as the destruction of the environment and the degradation of the robot also exist.

In contrast, methods based on human teaching such as teaching-playback~\cite{Playback_Lee2009, MotionReproduction_Horikoshi2024} and imitation learning~\cite{ImitationLearningSurvey_Hussein2017, SurveyImitationLearning_fang2019, OverviewImitationLearinig_Attia2018} have been proposed. These methods generate actions from human teaching, and therefore, they do not require explicit programming for motion generation. These methods are very efficient for obtaining trajectories required to perform a task and enable robots to act in complex motions.

Several methods for human instructions have been proposed. The simplest method is kinesthetic teaching~\cite{KinestheticTeaching_Amir2018, DynamicMovementPrimitives_gavspar2018}, which only outputs response values. Hence, it is difficult to reproduce high-speed movements. To address this issue, a method using a remote control to teach movements has been proposed~\cite{zhao_ALOHA, Fu_MobileALOHA}. Further, four-channel bilateral control~\cite{BilateObliquecordinate_Sakaino2011}, which is a remote control method that can teach both position and force information, has also been proposed.

Teaching-playback~\cite{Playback_Lee2009, MotionReproduction_Horikoshi2024} is a method where robots replicate movements using pre-recorded motion data. Motion-copying system~\cite{MotionCopyingStability_Yokokura2009, motioncopy_igarashi2015, MotionCopying_Fujisaki2023}, which is a type of teaching-playback, reproduces position and force data based on four-channel bilateral control. This can also reproduce force information, and therefore, it has a certain degree of adaptability to the environment. Robots can handle objects even if their shape change; however, it is difficult to adapt to large environmental changes because they only reproduce past actions. Further, it is difficult to change behavior in response to conditions. The advantages of this method are that it can reproduce motion with only one motion data and it is easy to predict motion because it follows the original motion.

Imitation learning~\cite{ImitationLearningSurvey_Hussein2017, SurveyImitationLearning_fang2019, OverviewImitationLearinig_Attia2018} is an end-to-end learning method for robots based on human instruction data. Methods that use Markov models~\cite{MarkovImitate_Asfour2008} and mixed Gaussian models~\cite{mixedGaussian_Silverio2018} have been proposed. Further, considerable research has been conducted on models using neural networks~\cite{DeepImitateLearn_Mochizuki2013, EndtoEnd_Levine2016, Virtual_Zhang2018, ImitateBilate_Adachi2018}. In imitation learning using deep learning, high adaptability to the environment is achieved by sequential motion planning. Bilateral control-based imitation learning, which utilizes four-channel bilateral control~\cite{ImitateBilate_Adachi2018, CooperationImitate_Sasagawa2020, ImitateBilate_Saigusa2022, BiACT_Buamanee2024, ILBiT_kobayashi2024} is very superior in that it can reproduce human force information. Moreover, it is possible to handle visual information in generating robot motions using visual information during motion as features of deep neural networks, and this is expected to improve the adaptive capability to the environment. Although multiple movements are required for learning, supervised learning is highly efficient, unlike reinforcement learning, wherein robots learn human instructions as correct movements.

In this study, we implemented and verified the motion-copying system and bilateral control-based imitation learning using a 19 degree-of-freedom (DOF) robot, Foodly TypeR. We participated in the Food Topping Challenge at ICRA 2024 and attempted serving salmon roe on rice and picking fried chicken. We won first place in the challenge of serving salmon roe on rice using the motion-copying system, and we won second place in the challenge of picking fried chicken using imitation learning.

This study demonstrates the current performance of motion generation methods based on human instruction. We described the implementation of the four-channel bilateral control, motion-copying system, and imitation learning in a dual-armed robot.

\section{Tasks}

\begin{figure}
\centering
\subfloat[Bowl for salmon roe.]{
    \resizebox*{4cm}{!}{
        \centering
        \includegraphics{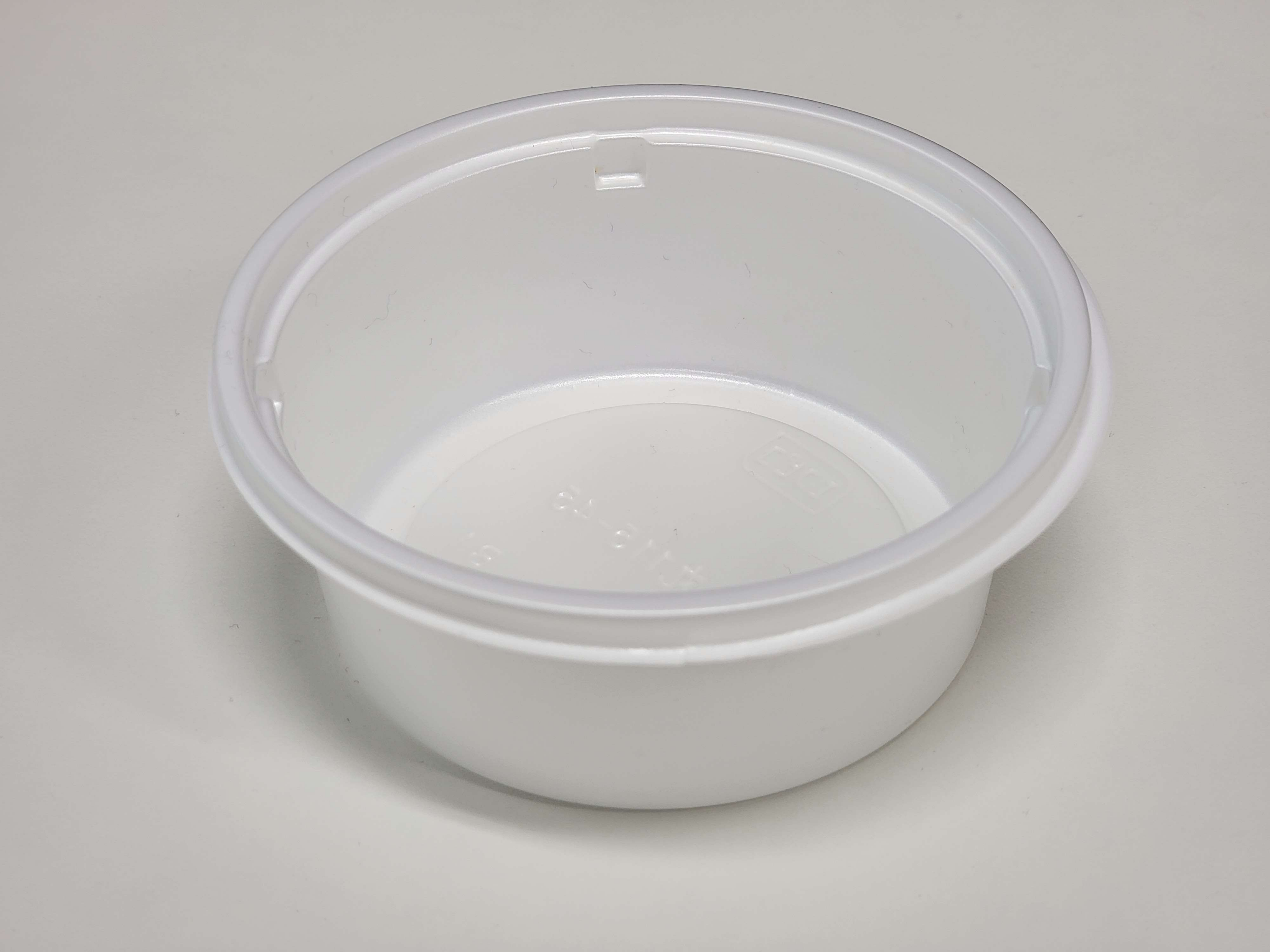}
        \label{fig:bowl1}}}
\hspace{20pt}
\subfloat[Container for fried chicken.]{
    \resizebox*{4cm}{!}{
        \centering
        \includegraphics{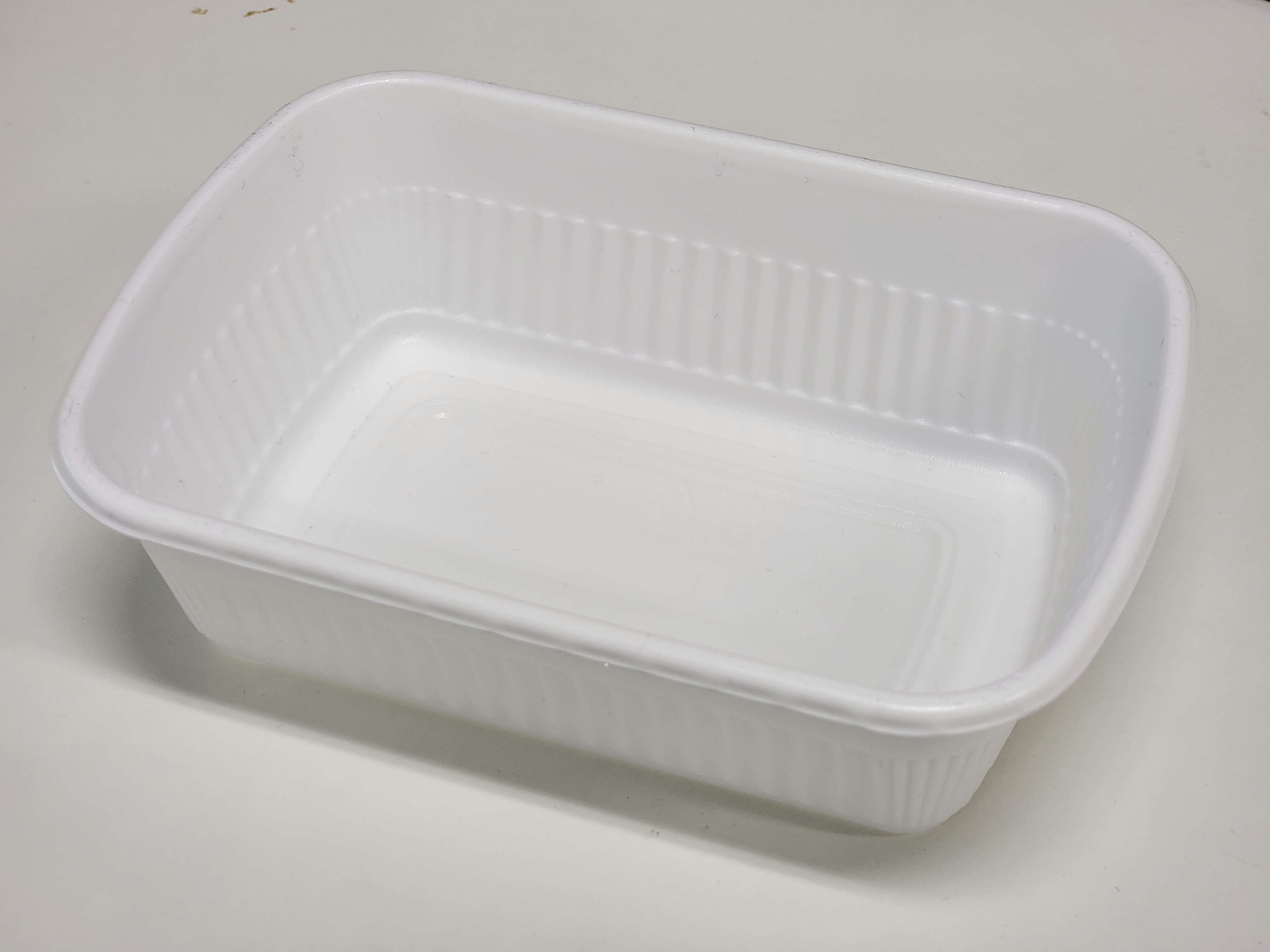}
        \label{fig:bowl2}}}
\hspace{20pt}
\subfloat[Fried chicken food sample.]{
    \resizebox*{4cm}{!}{
        \centering
        \includegraphics{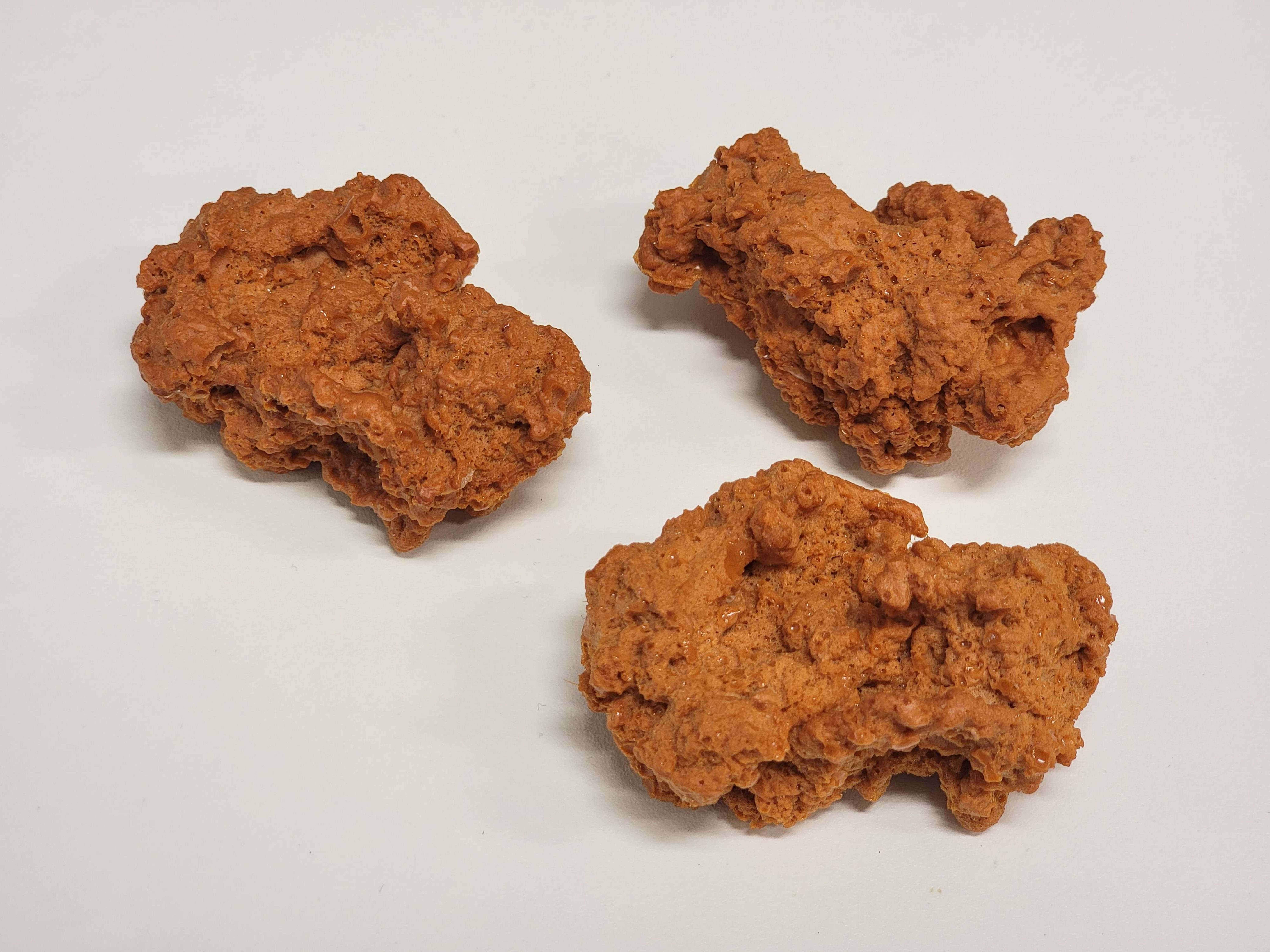}
        \label{fig:fired chicken}}}
\caption{Items for experiments.}
\label{fig:food_figs}
\end{figure}

The Food Topping challenge at ICRA 2024 required serving salmon roe on rice and picking fried chicken. Although there were no restrictions on the robot hardware, the use of Foodly Type-R was recommended, and therefore, we tackled these tasks using Foodly TypeR.

\subsection{Task 1: Serving Salmon Roe on Rice}

The serving task of salmon roe involved topping a rice bowl with salmon roe and pickled ginger. The bowl was made of polystyrene and had a diameter of 115~mm, height of 45~mm, and weight of 5~g. A photograph of the bowl is shown in Figure~\ref{fig:food_figs}\subref{fig:bowl1}. In addition, 100~g of rice was served in a bowl. The task was completed by placing the appropriate amount of salmon roe and pickled ginger in the rice bowl on the conveyor belt. The salmon roe was fully contained in the food tray and the proper amount had to be heaped. The rice and pickled ginger were real food, whereas the salmon roe was artificial food. Ten salmon roe bowls were served within a time limit of 20~minutes, and the completed salmon roe bowls were evaluated by the judges.

The evaluation criteria are
\begin{enumerate}
    \item Time required to serve ten dishes
    \item Variation in the weight of the ten dishes
    \item Appearance of the serving
    \item Amount of food loss
\end{enumerate}

There was no restriction on how to serve the salmon roe, and therefore, we scooped it up using a ladle. The salmon roe food tray was made of polypropylene and had a width of 540~mm, depth of 335~mm, height of 75~mm, and a capacity of 10.6~L. We prepared the appropriate amount of pickled ginger each time next to the food tray.

\subsection{Task 2: Picking Fried Chicken}

The task of picking fried chicken involved picking fried chicken randomly placed on a food tray and transferring it to a container placed on a conveyor belt. The container was made of polystyrene and had a width of 192~mm, depth of 133~mm, and height of 56~mm. A photograph of the container is shown in Figure~\ref{fig:food_figs}\subref{fig:bowl2}. The same food tray as in Task 1 was used. We used food samples for the fried chicken, which varied in size from 35-50~mm and in weight from 8-12~g. A photograph of the fried chicken is shown in Figure~\ref{fig:food_figs}\subref{fig:fired chicken}.

The evaluation criterion was the number of pieces picked and placed in 300~seconds. Grabbing multiple pieces at once was considered invalid, and points were deducted if a piece was dropped outside the container.

\section{Control System}
\subsection{Robot System}

\begin{figure}
\centering
\subfloat[Foodly TypeR.]{
    \resizebox*{4cm}{!}{
        \centering
        \includegraphics{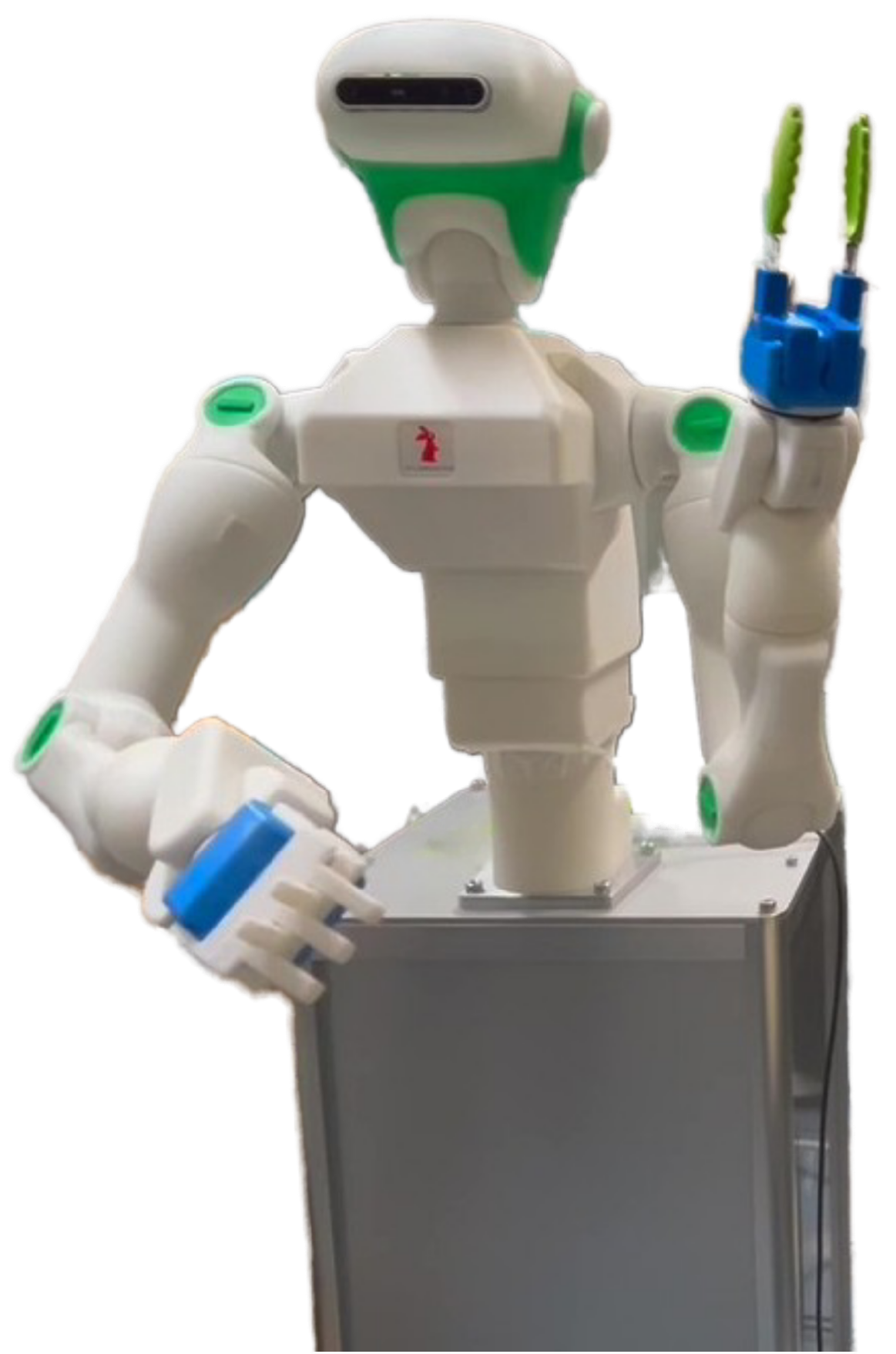}
        \label{fig:Foodly_TypeR}}}
\hspace{40pt}
\subfloat[Sciurus17.]{
    \resizebox*{4cm}{!}{
        \centering
        \includegraphics{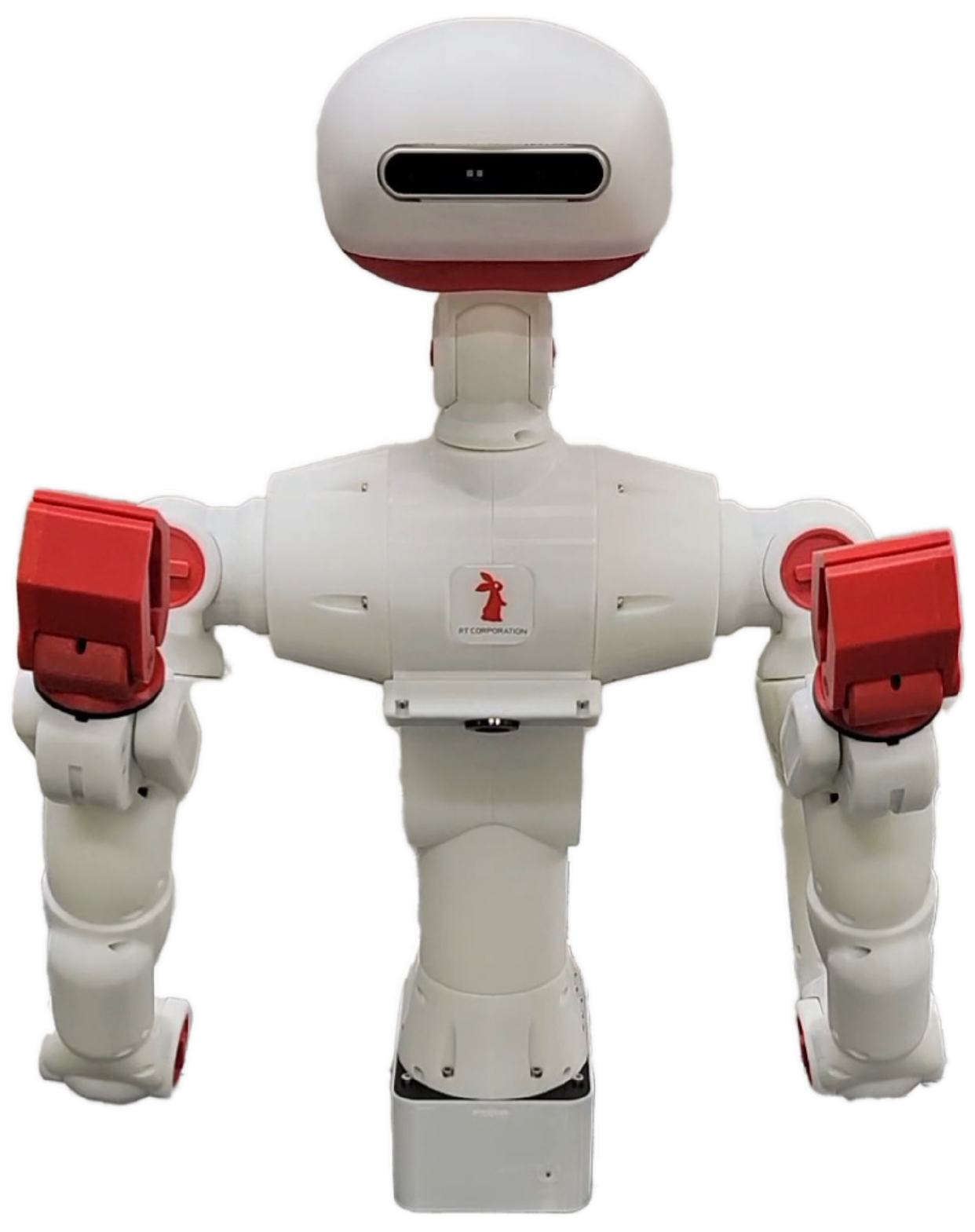}
        \label{fig:Sciurus17}}}
    \caption{Foodly TypeR is a humanoid collaborative robot manufactured by RT Corporation. The left hand is the default hand, and the right hand is the cross hand used in this experiment. Both hands were replaced in the experiment. Sciurus17 is a research-use upper-body humanoid robot manufactured by RT Corporation.}
    \label{fig:robots}
\end{figure}

We employed two 19-DOF robots, Foodly TypeR and Sciurus17, manufactured by RT Corporation. They have similar kinematics but different link lengths and weights. Foodly TypeR and Sciurus17 is shown in Figure~\ref{fig:robots}\subref{fig:Foodly_TypeR} and~\ref{fig:robots}\subref{fig:Sciurus17}, respectively. We replaced default hands with cross hands~\cite{GraspHand_Yamane2024} on both robots. Joint numbers were assigned in the order of right arm, left arm, and torso, starting from the root of the right arm. The right and left grippers were assigned 8 and 16 joints, respectively.

In this study, the robot was trained using four-channel bilateral control. A robot operated by a person is called leader, and a robot that performs the task is called follower. We used Sciurus17 as the leader and Foodly TypeR as the follower. Motion-copying system and autonomous operations of imitation learning were performed using Foodly TypeR.

Both robots were connected directly to an external desktop computer using an RS485 to USB converter for high-speed communication with the internal servo motors of the robot. Since the RS485 bus is present in the left arm, right arm, and torso, a total of six USB cables were connected to an external desktop computer. The intel RealSense Camera equipped with the head of Foodly TypeR was connected to the same desktop computer.

All servo motors were used in the current control mode, and torque reference values were transmitted from the desktop computer. The torque reference values and neural network inference were calculated using the desktop computer.

\subsection{Four-Channel Bilateral Control}

\begin{figure}
    \centering
    \includegraphics[scale=1.5]{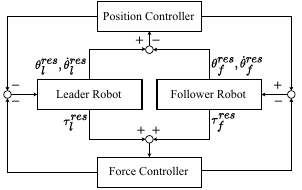}
    \caption{Block diagram of four-channel bilateral control.}
    \label{Fig:Block Diagram of Bilateral}
\end{figure}

Four-channel bilateral control is a teleoperation control method that synchronizes position and reaction force in both directions. It performs acceleration control and calculates acceleration command values from the position and reaction force response values. Acceleration control ensures that the difference in position between the leader and follower is zero and the sum of the reaction forces is zero. Position synchronization is achieved by setting the difference in position to zero. Further, the action-reaction law is realized by setting the sum of reaction forces to zero. These are respectively defined as;
\begin{equation}
    \label{Eq:Differential_Mode_of_Motion_ReTouch}
    \bm{\theta}_{l}^{res} = \bm{\theta}_{f}^{res},
\end{equation}
\begin{equation}
    \label{Eq:Common_Mode_of_Motion_ReTouch}
    \bm{\tau}_{l}^{res} + \bm{\tau}_{f}^{res} = \bm{0},
\end{equation}
where $\bm{\theta}^{res}$ and $\bm{\tau}^{res}$ represent angle and reaction force response vectors, respectively. The subscripts $\bigcirc_{l}$ and $\bigcirc_{f}$ represent the leader and the follower, respectively.

Figure~\ref{Fig:Block Diagram of Bilateral} shows the block diagram of the four-channel bilateral control. The controller and hardware parameters and the gravity compensator parameters are liested in Table~\ref{tab:parameters} and~\ref{tab:grav_parameters}, respectively.

{
\tabcolsep = 1.5pt
\begin{table}[]
    \centering
    \caption{Controller and hardware parameters}
    \label{tab:parameters}
    \begin{tabular}{lc|lllllll}
        \hline
        \begin{tabular}{l} Robot \\ Type \end{tabular} &
        \begin{tabular}{l} Joint \\ Number \end{tabular} &\,& 
        \begin{tabular}{l} Position \\ P gain \\ $K_{p\bigcirc}$ \end{tabular} &
        \begin{tabular}{l} Position \\ D gain \\ $K_{d\bigcirc}$ \end{tabular} &
        \begin{tabular}{l} Force \\ P gain \\ $K_{f\bigcirc}$ \end{tabular} &
        \begin{tabular}{l} Inertia \\ $J_\bigcirc$  \\ \lbrack kg m$^2$\rbrack \end{tabular} &
        \begin{tabular}{l} Viscous \\ Friction \\ Coefficient  \\ $D_\bigcirc$ \\ \lbrack N~m~s\rbrack \end{tabular} &
        \begin{tabular}{l} Cutoff \\ Frequency \\ $f_{C\bigcirc}$ \\ \lbrack rad\slash s\rbrack \end{tabular} \\ \hline\hline
\multirow{19}{*}{Sciurus17} & 1  & & 50  & 20  & 0.9 & 0.1099  & 0.4105 & 30\\ 
 & 2  & & 15  & 40  & 0.4 & 0.1147  & 0.7655 & 30\\ 
 & 3  & & 30  & 33  & 0.5 & 0.0432  & 0.2187 & 30\\ 
 & 4  & & 30  & 20  & 0.5 & 0.0582  & 0.2687 & 30\\ 
 & 5  & & 10  & 20  & 0.7 & 0.005676 & 0.0400  & 30\\ 
 & 6  & & 35  & 20  & 0.7 & 0.0066   & 0.0391 & 30\\ 
 & 7  & & 20  & 20  & 0.5 & 0.006281 & 0.0500  & 30 \\ 
 & 8  & & 80  & 20  & 0.9 & 0.006891 & 0.021 & 30 \\ 
 & 9  & & 50  & 20  & 0.9 & 0.1099  & 0.4105 & 30\\ 
 & 10 & & 15  & 40  & 0.4 & 0.1147  & 0.7655 & 30\\ 
 & 11 & & 30  & 33  & 0.5 & 0.0432  & 0.2187& 30\\ 
 & 12 & & 30  & 20  & 0.5 & 0.0582  & 0.2687 & 30\\ 
 & 13 & & 10  & 20  & 0.7 & 0.005676 & 0.0400  & 30\\ 
 & 14 & & 35  & 20  & 0.7 & 0.0066   & 0.0391 & 30\\ 
 & 15 & & 20  & 20  & 0.5 & 0.006281 & 0.0500 & 30  \\ 
 & 16 & & 80  & 20  & 0.9 & 0.006891 & 0.0210 & 30 \\ 
 & 17 & & 40  & 20  & 0.0 & 0.1000     & 0.0000 & 15   \\ 
 & 18 & & 128 & 20  & 0.0 & 0.0050   & 0.0000 & 20   \\ 
 & 19 & & 128 & 20  & 0.0 & 0.0050   & 0.0000 & 20   \\ 
\hline
\multirow{19}{*}{Foodly TypeR} & 1  & & 50  & 20  & 0.9 & 0.0649  & 0.1609 & 30 \\ 
 & 2  & & 15  & 40  & 0.4 & 0.1067  & 0.4466 & 30 \\ 
 & 3  & & 30  & 33  & 0.5 & 0.0536  & 0.2730 & 30 \\ 
 & 4  & & 30  & 40  & 0.5 & 0.0506  & 0.1706 & 30 \\ 
 & 5  & & 10  & 20  & 0.7 & 0.0100  & 0.0400 & 30 \\ 
 & 6  & & 35  & 20  & 0.7 & 0.0130  & 0.0391 & 30 \\ 
 & 7  & & 20  & 20  & 0.5 & 0.0120  & 0.0500 & 30 \\ 
 & 8  & & 150 & 20  & 0.9 & 0.0130  & 0.0210 & 30 \\ 
 & 9  & & 50  & 20  & 0.9 & 0.0649  & 0.1609 & 30 \\ 
 & 10 & & 15  & 40  & 0.4 & 0.1067  & 0.4466 & 30 \\ 
 & 11 & & 30  & 33  & 0.5 & 0.0536  & 0.2730 & 30 \\ 
 & 12 & & 30  & 40  & 0.5 & 0.0506  & 0.1706 & 30 \\ 
 & 13 & & 10  & 20  & 0.7 & 0.0100  & 0.0400 & 30 \\ 
 & 14 & & 35  & 20  & 0.7 & 0.0130  & 0.0391 & 30 \\ 
 & 15 & & 20  & 20  & 0.5 & 0.0120  & 0.0500 & 30 \\ 
 & 16 & & 150 & 20  & 0.9 & 0.0130  & 0.0210 & 30 \\ 
 & 17 & & 40  & 20  & 0.0 & 0.2000  & 0.0000 & 15 \\ 
 & 18 & & 128 & 20  & 0.0 & 0.0100  & 0.0000 & 20 \\ 
 & 19 & & 128 & 20  & 0.0 & 0.0100  & 0.0000 & 20 \\
    \end{tabular}
\end{table}}

\begin{table}
    \centering
    \caption{Gravity compensator parameters}
    \label{tab:grav_parameters}
    \begin{tabular}{lllll}
    \hline
    & Parameter & Robot Type & Value & Unit \\ \hline\hline
    \multirow{2}{*}{$m_{2}$} & \multirow{2}{*}{Mass of second link} & Sciurus17 & 0.1936 & kg \\
     &  & Foodly TypeR & 0.1958 & kg \\ \hline
    \multirow{2}{*}{$m_{3}$} & \multirow{2}{*}{Mass of third link} & Sciurus17 & 0.1936 & kg \\
     &  & Foodly TypeR & 0.1958 & kg \\ \hline
    \multirow{2}{*}{$m_{4}$} & \multirow{2}{*}{Mass of fourth link} & Sciurus17 & 0.3112 & kg \\
     &  & Foodly TypeR & 0.3203 & kg \\ \hline
    \multirow{2}{*}{$c_{2}$} & \multirow{2}{*}{Center of mass of second link} & Sciurus17 & 0.1936 & m \\
     &  & Foodly TypeR & 0.1958 & m \\ \hline
    \multirow{2}{*}{$c_{3}$} & \multirow{2}{*}{Center of mass of third link} & Sciurus17 & 0.1936 & m \\
     &  & Foodly TypeR & 0.1958 & m \\ \hline
    \multirow{2}{*}{$c_{4}$} & \multirow{2}{*}{Center of mass of fourth link} & Sciurus17 & 0.3948 & m \\
     &  & Foodly TypeR & 0.3965 & m \\ \hline
    \multirow{2}{*}{$l_{2}$} & \multirow{2}{*}{Length of second link} & Sciurus17 & 0.1903 & m \\
     &  & Foodly TypeR & 0.1933 & m \\ \hline
    \multirow{2}{*}{$l_{3}$} & \multirow{2}{*}{Length of third link} & Sciurus17 & 0.1870 & m \\
     &  & Foodly TypeR & 0.1909 & m \\
    \hline
    \end{tabular}
\end{table}

We assumed the dynamics of the manipulator as:
\begin{equation}
    \bm{J} \ddot{\bm{\theta}}^{res} = \bm{\tau}^{ref} - \bm{\tau}^{res} - \bm{D} \dot{\bm{\theta}}^{res} - \bm{g}
    \label{eq:motion}
\end{equation}
where $\bm{J}$, $\ddot{\bm{\theta}}^{res}$, $\bm{\tau}^{ref}$, $\bm{\tau}^{res}$, $\bm{D}$, and $\bm{g}$ represent the inertia matrix, response vector of the angular acceleration, control input vector, reaction force vector, viscous friction coefficient matrix, and gravity vector, respectively.

$\bm{J}$ and $\bm{D}$ are diagonal matrices expressed as :
\begin{equation}
    \bm{J} = \rm{diag}
    \begin{bmatrix}
        J_{1}, J_{2}, \cdots , J_{19}
    \end{bmatrix},
\end{equation}
\begin{equation}
    \bm{D} = \rm{diag}
    \begin{bmatrix}
        D_{1}, D_{2}, \cdots , D_{19}
    \end{bmatrix}.
\end{equation}

$\ddot{\bm{\theta}}^{res}$ is $\begin{bmatrix}\ddot{\theta}_{1}^{res}, \ddot{\theta}_{2}^{res}, \cdots ,\ddot{\theta}_{19}^{res}\end{bmatrix}^{T}$, where the subscript indicates the joint number. This is same for the angles, angler velocities, and reaction torques.

The gravity vector is expressed as:
\begin{equation}
    \bm{g} = 
    \begin{bmatrix}
        g_{1}, g_{2}, g_{3}, g_{4}, 0, 0, 0, 0, g_{9}, g_{10}, g_{11}, g_{12}, 0, 0, 0, 0, 0, 0, 0
    \end{bmatrix}^{T},
\end{equation}
where
\begin{equation}
    \begin{split}
    g_{1} &= g[-c_{2}m_{2}\sin{\theta_{1}}\sin{\theta_{2}} - m_{3}(c_{3} + l_{2})\sin{\theta_{1}}\sin{\theta_{2}} + m_{4}\{-c_{4}\sin{\theta_{4}}\cos{\theta_{1}}\cos{\theta_{3}} \\
    &-c_{4}\sin{\theta_{3}}\sin{\theta_{4}}\cos{\theta_{2}}\sin{\theta_{1}} - (-c_{4}\cos{\theta_{4}} + l_{2} + l_{3})\sin{\theta_{2}}\sin{\theta_{1}}\}], \\
    g_{2} &= g[c_{2}m_{2}\cos{\theta_{2}} + m_{3}(c_{3} + l_{2})\cos{\theta_{2}}
    + m_{4}\{-c_{4}\sin{\theta_{2}}\sin{\theta_{3}}\sin{\theta_{4}} \\
    &+ (-c_{4}\cos{\theta_{4}} + l_{2} + l_{3})\cos{\theta_{2}}\}]\cos{\theta_{1}}, \\
    g_{3} &= c_{4}gm_{4}(\sin{\theta_{1}}\sin{\theta_{3}} + \cos{\theta_{1}}\cos{\theta_{2}}\cos{\theta_{3}})\sin{\theta_{4}}, \\
    g_{4} &= c_{4}gm_{4}\{(\sin{\theta_{2}}\sin{\theta_{4}} + \sin{\theta_{3}}\cos{\theta_{2}}\cos{\theta_{4}})\cos{\theta_{1}} - \sin{\theta_{1}}\cos{\theta_{3}}\cos{\theta_{4}}\}, \\
    g_{9} &= g[c_{2}m_{2}\sin{\theta_{9}}\sin{\theta_{10}} + m_{3}(c_{3} + l_{2})\sin{\theta_{9}}\sin{\theta_{10}} - m_{4}\{ c_{4}\sin{\theta_{12}}\cos{\theta_{11}}\cos{\theta_{9}} \\
    &+ c_{4}\sin{\theta_{11}}\sin{\theta_{12}}\cos{\theta_{10}}\sin{\theta_{9}} - (-c_{4}\cos{\theta_{12}} + l_{2} + l_{3})\sin{\theta_{10}}\sin{\theta_{9}} \} ],\\
    g_{10} &= -g[c_{2}m_{2}\cos{\theta_{10}} + m_{3}(c_{3} + l_{2})\cos{\theta_{10}} + m_{4} \{ c_{4}\sin{\theta_{11}}\sin{\theta_{12}}\sin{\theta_{10}} \\
    &+ (-c_{4}\cos{\theta_{12}} + l_{2} + l_{3})\cos{\theta_{10}} \} ]\cos{\theta_{9}},\\
    g_{11} &= c_{4}gm_{4}(\sin{\theta_{11}}\sin{\theta_{9}} + \cos{\theta_{11}}\cos{\theta_{9}}\cos{\theta_{10}})\sin{\theta_{12}},\\
    g_{12} &= c_{4}gm_{4}\{(\sin{\theta_{11}}\cos{\theta_{12}}\cos{\theta_{10}} - \sin{\theta_{12}}\sin{\theta_{10}})\cos{\theta_{9}} - \sin{\theta_{9}}\cos{\theta_{11}}\cos{\theta_{12}}\},
    \end{split}
\end{equation}
where $g$, $m_{\bigcirc}$, $c_{\bigcirc}$, and $l_{\bigcirc}$ represent a gravity acceleration, weight of the link, distance from the joint to the center of mass of the link, and length of the link, respectively.

\begin{figure}
    \centering
    \includegraphics[scale=1.2]{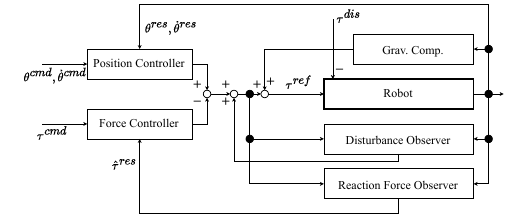}
    \caption{Block diagram of controller.}
    \label{Fig:Block Diagram of Controller}
\end{figure}

Figure~\ref{Fig:Block Diagram of Controller} shows the block diagram of the controller, where the disturbance is represented by $\bm{\tau}^{dis}$. The acceleration control is realized using gravity compensation and a disturbance observer (DOB)~\cite{DOB_Ohnishi1996, DOB35_Sariyildiz2020}. The sum of gravity compensation and output of DOB is represented by $\hat{\bm{\tau}}^{dis}$; the reaction force is estimated by a reaction force observer (RFOB)~\cite{Murakami_RFOB}. The estimated reaction force is represented by $\hat{\bm{\tau}}^{res}$. Moreover, angular velocities are calculated by pseudo-differentiation using the values obtained from the angle sensor.

In the position controller and force controllers, we used a proportional-derivative (PD) controller of position and a proportional (P) controller of force, respectively. The gain of the PD and P controller, $\bm{K}_{p}$, $\bm{K}_{d}$, and $\bm{K}_{f}$, were determined manually and expressed as:
\begin{equation}
    \bm{K}_{p} = \rm{diag}
    \begin{bmatrix}
        K_{p1}, K_{p2}, \cdots , K_{p19}
    \end{bmatrix},
\end{equation}
\begin{equation}
    \bm{K}_{d} = \rm{diag}
    \begin{bmatrix}
        K_{d1}, K_{d2}, \cdots , K_{d19}
    \end{bmatrix},
\end{equation}
\begin{equation}
    \bm{K}_{f} = \rm{diag}
    \begin{bmatrix}
        K_{f1}, K_{f2}, \cdots , K_{f19}
    \end{bmatrix}.
\end{equation}

The following equation provides the torque references for the leader and follower.
\begin{equation}
    \begin{split}
        \bm{\tau}_{l}^{ref} &=
         \frac{\bm{J}}{2} (\bm{K}_{p} + s \bm{K}_{d})(\bm{\theta}_{f}^{res} -\bm{\theta}_{l}^{res}) \\
         &- \frac{1}{2} \bm{K}_f (\hat{\bm{\tau}}_{f}^{res} + \hat{\bm{\tau}}_{l}^{res}) + \hat{\bm{\tau}}_{l}^{dis}, \\ 
        \bm{\tau}_{f}^{ref} &=
         \frac{\bm{J}}{2} (\bm{K}_{p} + s \bm{K}_{d})(\bm{\theta}_{l}^{res} -\bm{\theta}_{f}^{res}) \\
         &- \frac{1}{2} \bm{K}_f (\hat{\bm{\tau}}_{l}^{res} + \hat{\bm{\tau}}_{f}^{res}) + \hat{\bm{\tau}}_{f}^{dis},
    \end{split}
\end{equation}
where $s$ represents a Laplace operator.

The control cycle was set to 500~Hz. The cut-off frequencies of the low-pass filters of the DOB, RFOB, and pseudo-differential were the same, and their values were $f_{C1}, f_{C2}, \cdots, f_{C19}$.

\subsubsection{System Identification}

Determining the physical parameters of robots is necessary to implement four-channel bilateral control. To this end, system identification is performed as indicated below.

For the system identification phase, we employed the teleoperation of position using two Sciurus17 units because we had two Sciurus17 units available. However, the system can be identified in the same manner by using Foodly TypeR as the leader and Sciurus17 as the follower.

Using teleoperation, we moved the leader randomly as long as the follower did not touch an external object. The current command values input to the follower and the angle values were recorded at 500~Hz.

Simultaneously exciting all joints is difficult, and therefore, 10~minutes of data was collected for joints 1 to 4 and 5 to 8. Futher, ten~seconds of data was extracted as test data and used to evaluate the effectiveness of the parameters instead of updating them. We used sequential least squares to optimize the inverse dynamics calculation parameters. We excluded the errors of non-moving joints from the overall error because the dynamics may not follow equations~\ref{eq:motion} owing to the influence of static friction forces~\cite{SystemIdentification_Inami2024}.

Similarly, we identified the physical parameters of Foodly TypeR.

Motion data without angular limits or touching external objects for system identification can be obtained easily using remote control. In addition, systems with motion ranges and speeds close to those of actual use can be identified based on random human motions.

\subsection{Motion-Copying System (Serving Salmon Roe on Rice)}
\subsubsection{Data Collection}

\begin{figure}
    \centering
    \includegraphics{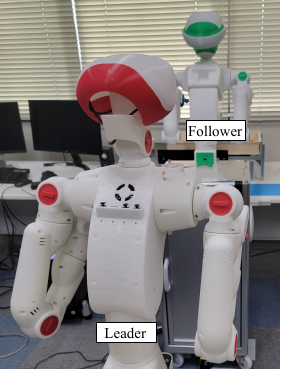}
    \caption{Positioning for teaching movements.}
    \label{Fig:Leader_behind_follower}
\end{figure}
We obtained the response values of the leader and follower using four-channel bilateral control. Motion data for the right and left arms were collected simultaneously. The leader was placed behind the follower as shown in Figure~\ref{Fig:Leader_behind_follower}, and the motion was taught while the behavior of the follower could be checked. Right-handed and left-handed persons were in charge of teaching the right and left arms, respectively.

The sampling frequency of the four-channel bilateral control was 500~Hz, and motion records of the same timing were maintained. The movement data was sped up by twice the speed to achieve faster autonomous movements, and motion-copying was performed based on the double-speed data by skipping over each line of the time series data and doubling only the angular velocity. Doubling only motion data amplified the noise and made the motion unstable. Therefore, a zero-phase filter was applied to all angular, angular velocity, and reaction force data to reduce noise. The coefficients of the zero-phase filter are the same as $f_{C1}, f_{C2}, \cdots, f_{C19}$. Teaching at low speeds helps achieve complex motions. However, steep motions can be omitted because of the filtering. It was important to perform teaching based on this characteristic.

\begin{figure}
    \centering
    \includegraphics[scale=1.5]{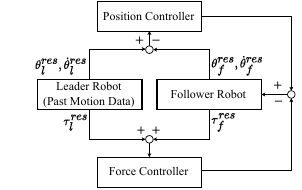}
    \caption{Block Diagram of Motion-copying System.}
    \label{Fig:Block Diagram of Motion Copying}
\end{figure}

\subsubsection{System}
The motion-copying system can reproduce a motion obtained using four-channel bilateral control. The response values of the leader angle, angular velocity, and reaction force values are used as command values to achieve autonomous motion. The block diagram of the motion-copying system is shown in Figure~\ref{Fig:Block Diagram of Motion Copying}. The controller is the same as that of the four-channel bilateral control.

The motion-copying system is based on hybrid position/force control, and therefore, it can respond to a certain degree to environmental fluctuations. However, this does not change command values, and therefore, the system cannot respond to large trajectory changes or large-scale environmental fluctuations.

\subsection{Imitation Learning (Picking Fried Chicken)}
\subsubsection{Data Collection}

We obtained motion data using the same procedure used for the motion-copying system. In addition, we collected 35~sequences of data each for the right and left arms. We used 32~sequences as training data and three sequences as validation data.

The sampling frequency of the four-channel bilateral control was 500~Hz, and motion records of the same timing were retained. Data augmentation was performed by downsampling at 20~ms and shifting the starting point of the data by 2~ms~\cite{MultipleData_Rahmatizadeh2018}, which expanded the data by a factor of 10. We normalized the data to have a mean of zero and a standard deviation of one. In addition, white noise with a variance of 0.01 was added to the normalized data.

\subsubsection{Model and System}

\begin{figure}
    \centering
    \includegraphics[scale=1.5]{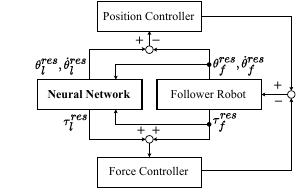}
    \caption{Block Diagram of Four-channel Bilateral Control based Imitation Learning.}
    \label{Fig:Block Diagram of imitation learning}
\end{figure}
 
We employed neural networks to generate adaptive behavior to changes in object location.

In bilateral control-based imitation learning, motion generation is achieved by predicting the command of the follower based on its response. Since the command of the follower is the response of the leader, the neural network can be trained by predicting the response of the leader from the response of the follower when learning. A block diagram of imitation learning is shown in Figure~\ref{Fig:Block Diagram of imitation learning}. The controller of the robot is the same as that of the four-channel bilateral control.

\begin{figure}
    \centering
    \includegraphics[width=0.95\linewidth]{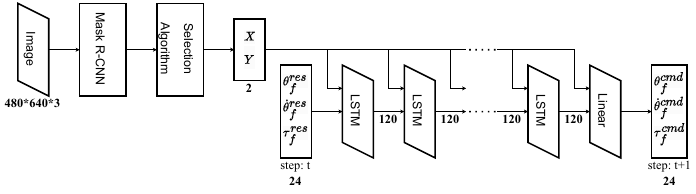}
    \caption{Neural Network Architecture.}
    \label{Fig:Neural Network Architecture}
\end{figure}

The architecture of the neural network is shown in Figure~\ref{Fig:Neural Network Architecture}. The center coordinates of fried chicken are extracted using Mask R-CNN~\cite{MaskRCNN_He2017} from the RGB image pictured by Realsense on the head of Foodly TypeR. When there are multiple fried chicken in the image, all distances between the centers of the fried chicken in the image are calculated, and the object with the largest minimum distance from other fried chicken is selected to avoid picking multiple objects simultaneously.

The Mask R-CNN model was the same as that in ref.~\cite{MaskRCNN_He2017}. Stochastic gradient descent with a learning rate of 0.005, momentum of 0.9, and coefficient of L2 regularization of 0.0005 was used for parameter optimization. We prepared ten images with 30 pieces of fried chicken and used nine images as training data and one image as test data. Before each training session, we randomly applied a color jitter, a flip, and an affine transformation. In the color jitter, the brightness, contrast, and saturation were varied by values obtained from a uniform distribution ranging from 0.7-1.3, respectively. In the flip, the images were flipped vertically and horizontally, each with a probability of 50\%. In the affine transform, images were randomly rotated in the zero to one-degree range, and the image was randomly shifted vertically and horizontally, each with a maximum of 10\%.

Neural networks were independently applied to the left and right arms to generate motion. The neural network was an 8-layer 120-dimensional long short-term memory (LSTM)~\cite{LSTM_Hochreiter1997}. The features of LSTM inclueded the coordinates of the object and response values of the angle, angular velocity, and reaction force of the follower. The coordinates in the image were the X and Y coordinates and added to each layer of the LSTM~\cite{EachLayer_Yamane2024}. The objective variables were the command values of the angle, angular velocity, and reaction force of the follower after one step. We used the Adam optimizer~\cite{Adam_kingma2014} with a learning rate of 0.001 for parameter optimization. 

In Foodly TypeR, the right and left arms often tangled when attempting to simultaneously grasp fried chicken near the center of the tray. To address this issue, an interference band was added near the center to prevent fried chicken in this area from being picked. In addition, fried chicken near the edges of the tray was excluded from being picked to avoid issues where the tray edge obstructed the grasp.

Since the neural network operates at 50~Hz, the command values were updated once every 20~ms. From the point of view of the robot controller, the command values change once every 10 times, and in other cases, the previous command values are maintained.

\section{Results and Discussions}

\subsection{Task1: Serving Salmon Roe on Rice (Motion-Copying System)}

\begin{figure}
  \begin{minipage}[b]{0.4\linewidth}
    \centering
    \includegraphics[width=0.8\linewidth]{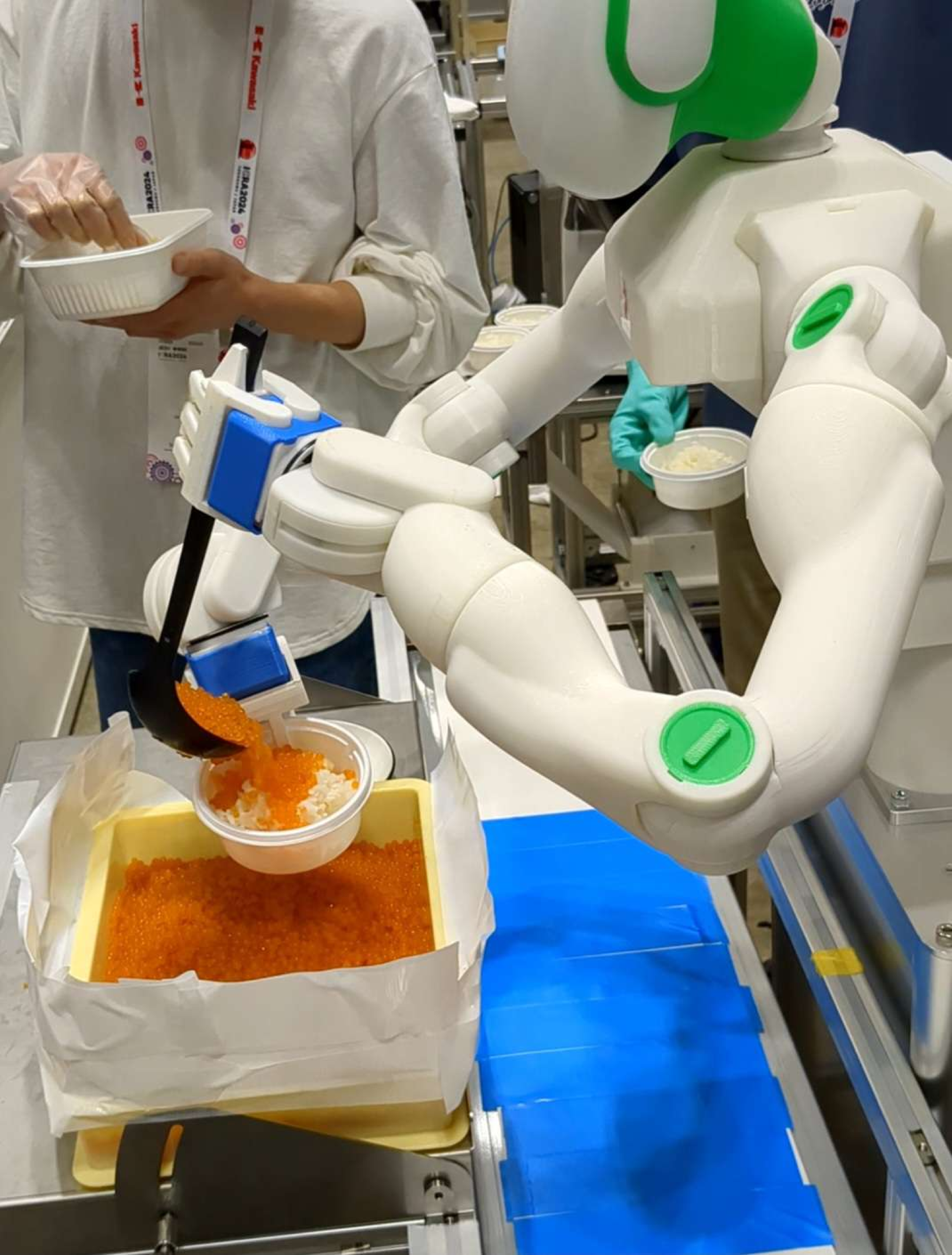}
    \caption{Serving Salmon Roe on Rice}
    \label{fig:ikraing}
  \end{minipage}
  \begin{minipage}[b]{0.5\linewidth}
    \centering
    \includegraphics[width=0.8\linewidth]{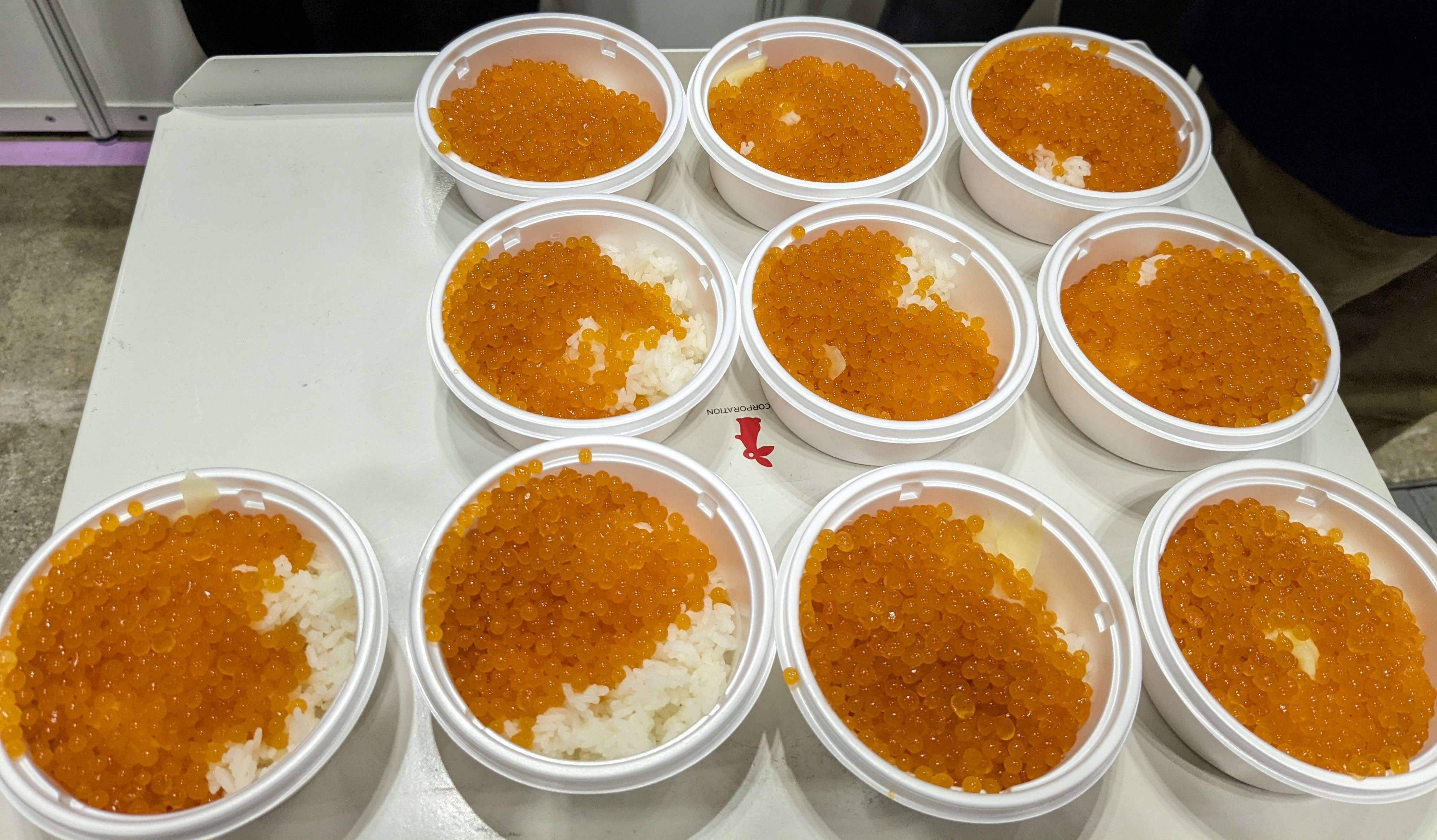}
    \caption{Finished salmon roe bowls.}
    \label{fig:ikra}
  \end{minipage}
\end{figure}

The situation during the operation of serving salmon roe bowls is shown in Figure~\ref{fig:ikraing}, and the completed salmon roe bowls are shown in Figure~\ref{fig:ikra}.

The time required to serve 10 salmon roe bowls was 257~seconds; this was the fastest time among all participating teams. The weight standard deviation was 7\%, which was also the smallest among all participating teams. Further, the food loss was also minimal.

The robot was taught to move in approximately 40~seconds, and by doubling the speed, an autonomous movement of approximately 20~seconds was achieved. In addition, it took approximately 5~seconds to start the next operation after the operation was completed.

We made several ingenuities in teaching the movements. First, we smoothed the surface of the salmon roe with the back of the ladle before scooping, which reduced the variation in weight. Second, we served the salmon roe at the top of the food tray, which prevented loss if some fell out. Although these movements come naturally to humans, they are challenging to implement in a robot.

\subsection{Task2: Picking Fried Chicken (Imitation Learning)}

\begin{figure}
  \begin{minipage}[b]{0.5\linewidth}
    \centering
    \includegraphics[width=0.8\linewidth]{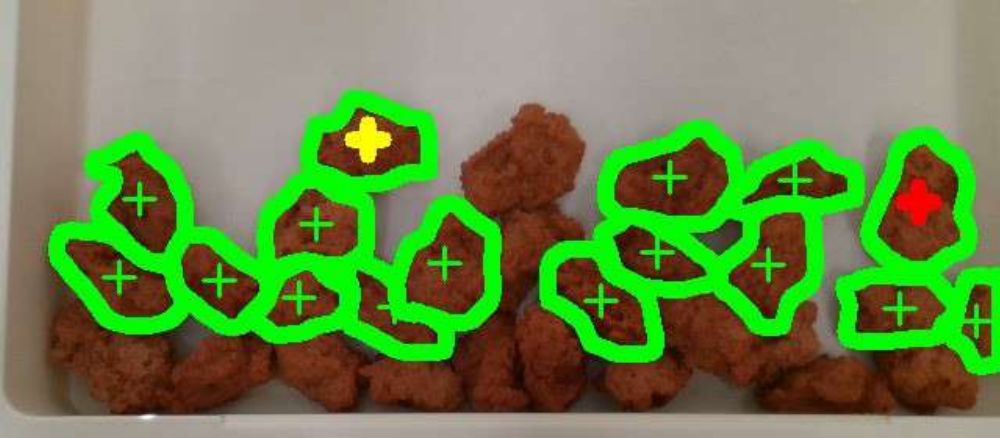}
    \caption{Recognition using Mask R-CNN.}
    \label{Fig:MaskRCNN}
  \end{minipage}
  \begin{minipage}[b]{0.4\linewidth}
    \centering
    \includegraphics[width=0.8\linewidth]{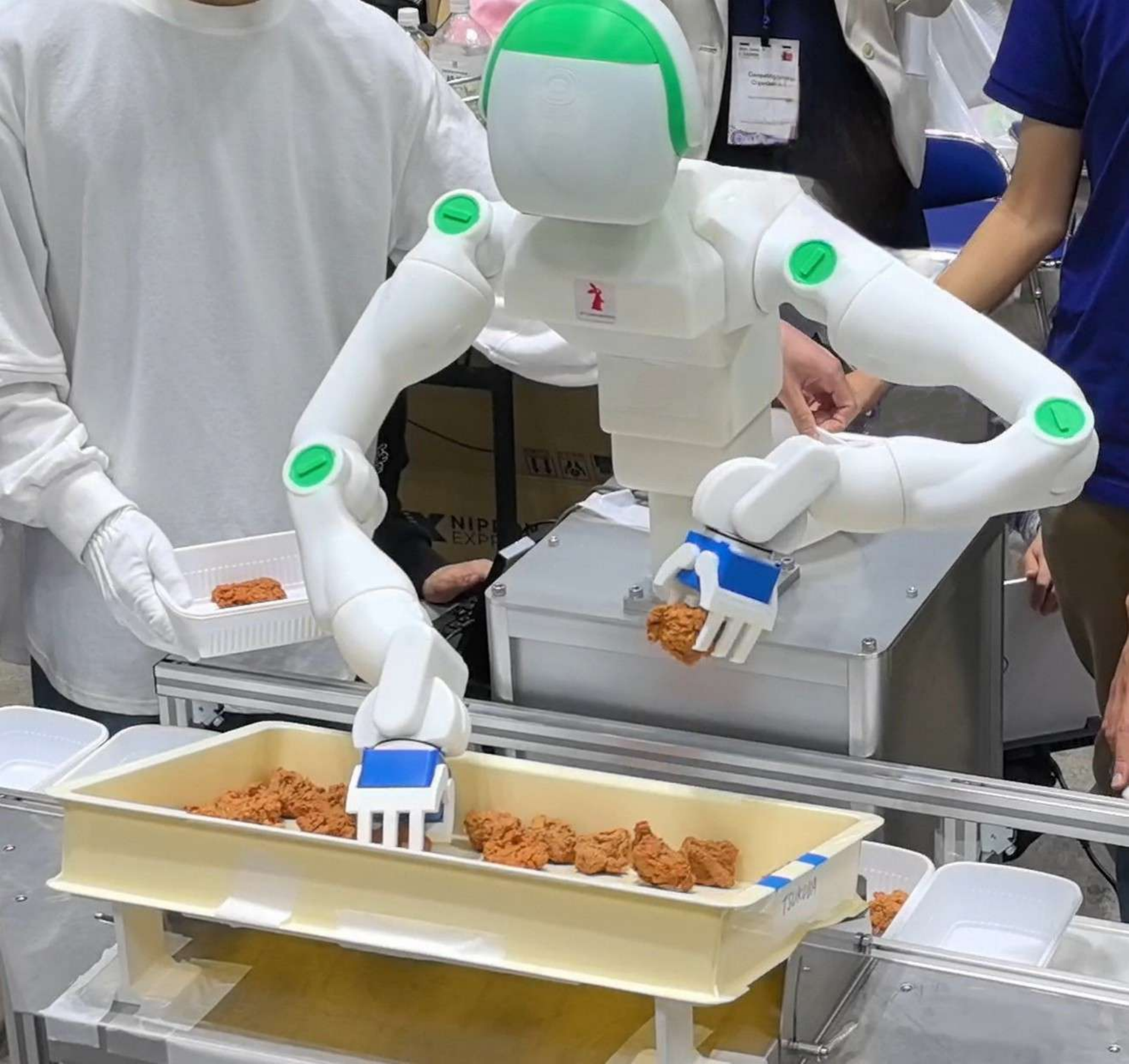}
    \caption{Picking Fried Chicken.}
    \label{Fig:picking}
  \end{minipage}
\end{figure}

The result of the Mask R-CNN validation is shown in Figure~\ref{Fig:MaskRCNN}. The center and edges of the food tray are ignored; fried chicken picked by the right and left hands are marked in red and yellow, respectively; and the farthest piece from others is also highlighted.

The situation when picking the fried chicken is shown in Figure~\ref{Fig:picking}.

We successfully picked and placed 26 pieces of fried chicken in 300~seconds. The number of successful pickings was the highest among the participating teams. However, we accidentally picked two pieces of fried chicken three times and dropped one piece. In addition, we also failed to grab a piece of fried chicken 16 times. Consequently, we were runners-up in the comprehensive evaluation.

In this experiment, owing to the use of the cross hand, we sometimes grabbed several pieces of fried chicken simultaneously when multiple pieces were nearby. The selection algorithm aimed to pick the piece furthest from others; however, in some cases, adjacent pieces were picked. In addition, the cross hand occasionally pierced the chicken, preventing proper release and causing it to fall.

\section{Conclusion}

We used Foodly TypeR to take on the ICRA 2024 Food Topping Challenge, using four-channel bilateral control to teach human motion. We achieved fast and flexible motion by simultaneously providing both position and force information to the robot. Using the motion-copying system, the robot replayed the human-taught motion and completed the task of serving salmon roe on rice. Consequently, we successfully served 10 salmon roe bowls the fastest among all participating teams, minimizing weight variance and food waste. Further, using bilateral control-based imitation learning, the robot adapted its pick-and-place motion to the position of the fried chicken. Our team achieved the highest pick-and-place count within the time limit. However, we lost one point for dropping a piece of fried chicken, placing us second in the final evaluation.

The robot often failed to grasp fried chicken; and therefore, recovery from such failures must be achieved in future~\cite{Recovery_Takeuchi2023}. Moreover, the high control performance of four-channel bilateral control is essential for more accurate and faster motion teaching, which requires highly precise acceleration control and RFOB~\cite{Jerk_Nagao2023, HighRFOB_Phuong2023}.

Our nine-member team implemented four-channel bilateral control, motion-copying system, and imitation learning in about six weeks. We took a month to establish the four-channel bilateral control, while the motion teaching and generation phases were completed in about two weeks. Since the four-channel bilateral control has been constructed, we expect to teach Foodly TypeR new actions in a shorter time in future.

\section*{Disclosure statement}

  The authors declare no potential conflicts of interest.

\section*{Funding}

This work was supported by JSPS KAKENHI Grant Number 24K00905, JST, PRESTO Grant Number JPMJPR24T3 Japan and JST ALCA-Next Japan, Grant Number JPMJAN24F1. This study was based on the results obtained from the JPNP20004 project subsidized by the New Energy and Industrial Technology Development Organization (NEDO).

\section*{Notes on contributor(s)}

    Koki Inami received the degree of Policy and Planning Sciences in policy and planning sciences from University of Tsukuba, Japan, in 2024. He is currently working on an M.E. degree in intelligent and mechanical interaction systems at the Graduate School of Science and Technology at University of Tsukuba. His research interests include motion control, robotics, artificial intelligence, and machine learning. 

    Masashi Konosu received the B.E. degree in engineering systems from University of Tsukuba, Japan, in 2024. He is currently working on an M.E. degree in intelligent and mechanical interaction systems at the Graduate School of Science and Technology at University of Tsukuba. His research interests include motion control, robotics, artificial intelligence, and machine learning. 

    Koki Yamane received the B.E. degree in engineering systems and M.E. degree in intelligent and mechanical interaction systems from University of Tsukuba, Japan, in 2022 and 2024, respectively. He is currently pursuing a Ph.D. in intelligent and mechanical interaction systems at University of Tsukuba, Japan. His research interests include motion control, robotics, image processing, and machine learning.

    Nozomu Masuya received the B.E. degree in engineering systems from University of Tsukuba, Japan, in 2023. He is currently working on an M.E. degree in intelligent and mechanical interaction systems at the Graduate School of Science and Technology at University of Tsukuba. His research interests include motion control, robotics, artificial intelligence, and machine learning. 

    Yunhan Li received the B.E. degree in electronic information engineering (leeds) from Southwest Jiaotong University, China, in 2022. He is currently working on an M.E. degree in intelligent and mechanical interaction systems at the Graduate School of Science and Technology at University of Tsukuba. His research interests include motion control, robotics, artificial intelligence, and machine learning.

    Yu-Han Shu received the B.S. degree in department of electrical engineering from National Chi Nan University, Taiwan, in 2023. She is currently working on an M.E. degree in intelligent and mechanical interaction systems at the Graduate School of Science and Technology at University of Tsukuba. Her research interests include motion control, robotics, artificial intelligence, and machine learning.

    Hiroshi Sato received the B.E. degree from Shibaura Institute of Technology, Tokyo, Japan, in 2023. He is currently working on an M.E. degree in intelligent and mechanical interaction systems at the Graduate School of Science and Technology at University of Tsukuba, Japan. His research interests include motion control, robotics, and machine learning.

    Shinnosuke Homma received the B.E. degree in faculty of engineering from Saitama University, Japan, in 2024. He is currently working on an M.E. degree in intelligent and mechanical interaction systems at the Graduate School of Science and Technology at University of Tsukuba. His research interests include motion control, robotics, artificial intelligence, and machine learning. 

    Sho Sakaino received the B.E. degree in system design engineering and the M.E. and Ph.D. degrees in integrated design engineering from Keio University, Yokohama, Japan, in 2006, 2008, and 2011, respectively. He was an assistant professor at Saitama University between 2011 and 2019. Since 2019, he has been an associate professor at University of Tsukuba. His research interests include mechatronics, motion control, robotics, and haptics. He received the IEEE IES Best Conference Paper Award in 2022. He also received the IEEJ Industry Application Society Distinguished Transaction Paper Award in 2011 and 2020 and the RSJ Advanced Robotics Excellent Paper Award in 2020.

\bibliographystyle{tfnlm}
\bibliography{reference}

\end{document}